\documentclass[conference]{IEEEtran}
\usepackage{comment}
\IEEEoverridecommandlockouts
\usepackage{tikz}
\usepackage[caption=false,font=normalsize,labelfont=sf,textfont=sf]{subfig}
\usetikzlibrary{positioning, arrows, calc}
\usepackage{cite}
\usepackage{amsmath,amssymb,amsfonts}
\usepackage{algorithmic}
\usetikzlibrary{arrows.meta, positioning, shapes.geometric}
\usepackage{graphicx}
\usepackage{float}
\usepackage{stfloats}
\usepackage{textcomp}
\usepackage{xcolor}
\def\BibTeX{{\rm B\kern-.05em{\sc i\kern-.025em b}\kern-.08em
    T\kern-.1667em\lower.7ex\hbox{E}\kern-.125emX}}
\begin{document}

\title{Integration of Computer Vision with Adaptive Control for Autonomous Driving Using ADORE}
\author{
\IEEEauthorblockN{1\textsuperscript{st} Dr. Abu Shad Ahammed}
\IEEEauthorblockA{\textit{Chair of Embedded Systems} \\
\textit{University of Siegen}\\
Siegen, Germany \\
abu.ahammed@uni-siegen.de}
\and
\IEEEauthorblockN{2\textsuperscript{nd} Md Shahi Amran Hossain}
\IEEEauthorblockA{\textit{Chair of Embedded Systems} \\
\textit{University of Siegen}\\
Siegen, Germany \\
Md.Hossain@uni-siegen.de}
\and
\IEEEauthorblockN{3\textsuperscript{rd} Sayeri Mukherjee}
\IEEEauthorblockA{\textit{Chair of Embedded Systems} \\
\textit{University of Siegen}\\
Siegen, Germany \\
sayeri.mukherjee@student.uni-siegen.de}
\and
\IEEEauthorblockN{4\textsuperscript{th} Prof. Dr. Roman Obermaisser}
\IEEEauthorblockA{\textit{Chair of Embedded Systems} \\
\textit{University of Siegen}\\
Siegen, Germany \\
roman.obermaisser@uni-siegen.de}
\and
\IEEEauthorblockN{5\textsuperscript{th} Md. Ziaur Rahman}
\IEEEauthorblockA{\textit{Department of Computer Science and Engineering} \\
\textit{United International University}\\
Dhaka, Bangladesh \\
zia.rahman14@yahoo.com}
}

\maketitle

\begin{center}
    \textbf{\textit{This work has been submitted to the IEEE for possible publication. Copyright may be transferred without notice, after which this version may no longer be accessible.}}
\end{center}
\begin{abstract}
Ensuring safety in autonomous driving requires a seamless integration of perception and decision making under uncertain conditions. Although computer vision (CV) models such as YOLO achieve high accuracy in detecting traffic signs and obstacles, their performance degrades in drift scenarios caused by weather variations or unseen objects. This work presents a simulated autonomous driving system that combines a context-aware CV model with adaptive control using the ADORE framework. The CARLA simulator was integrated with ADORE via the ROS bridge, allowing real-time communication between perception, decision, and control modules. A simulated test case was designed in both clear and drift weather conditions to demonstrate the robust detection performance of the perception model while ADORE successfully adapted vehicle behavior to speed limits and obstacles with low response latency. The findings highlight the potential of coupling deep learning–based perception with rule-based adaptive decision making to improve automotive safety critical system.
\end{abstract}
\section{Introduction}
Artificial intelligence (AI) is a trending technology applied in different knowledge sections, including the automotive sector. Computer vision (CV), a part of AI, is now showing high potential on developing a safety critical system for autonomous driving. However, several challenges remain before adapting CV in real-world autonomy, as the performance of the model varies a lot when there are drifted scenarios like foggy weather, unknown objects which can endanger lives of drivers and other road users. In our previous research \cite{ahammed2025cv, hossain2024impact}, we already described such scenarios, and mitigation was proposed through drift training of the CV models, i.e. YOLO. The models developed to detect road signs and obstacles in the construction zone in drift scenarios achieved more than 90\% accuracy in validation and test. However, achieving good detection performance through perception models alone does not guarantee safe autonomous driving. There remains still a significant gap in our understanding of how computer vision models operate when employed in real time within a vehicle system. Traditional AI perception models often operate in isolation, without a direct feedback loop to the vehicle's control system. This disconnect can lead to delayed or suboptimal driving decisions, especially in safety-critical situations. Moreover, contemporary automotive research is not focused solely on replacing human drivers with autonomous systems via CV. Instead, it aims to explore the wider possibilities of autonomous driving by ensuring interaction with other road users and infrastructure. To facilitate such extensive functionality of autonomous driving, the German Aerospace Center (DLR) developed an open-source software framework ADORE (Automated Driving Open Research Environment) which provides a reference architecture and tools for implementing and evaluating driving behavior in simulation and, eventually, real-world scenarios.\\
In this automotive research, we have used the open source automotive simulator Car Learning to ACT (CARLA) to simulate an automotive environment where ADORE served as the control backbone, receiving input from external perception modules and making driving decisions accordingly. Although CARLA is capable of providing high-fidelity automotive simulation, photorealistic environments, and custom road mapping, it lacks a built-in, standardized decision-making and behavioral planning system suitable for advanced modular research. ADORE fills this gap by offering a flexible and semantically structured planning framework that integrates natively with ROS and supports real-time decision making. To the best of our knowledge, ADORE is currently the only open-source modular planning system compatible with CARLA for full-stack autonomy research. The integration enables the ego vehicle in CARLA's simulation to be managed for comprehensive evaluation of the full driving process, spanning from sensing to actuation, across diverse traffic conditions. The goal of this integration was two-fold: to examine how, at runtime, the computer vision model processes road instruments under drift conditions, and whether ADORE can adapt autonomous behavior accordingly, such as slowing down or speeding up for detected traffic signs, or stopping the vehicle if needed. This integrated approach bridges the gap between recognition and adaptive decision-making and enables the autonomous system to operate reliably in a wider range of scenarios.\\
The rest of the manuscript is organized as follows. Section II reviews related work and summarizes key developments in autonomous driving and safety-critical perception systems. Section III introduces the context-aware computer vision model and discusses its role in handling data drift. Section IV explains the integration of CARLA and ADORE through the ROS bridge. Section V presents the overall system workflow, while Section VI describes the verification experiments and evaluates system performance. Finally, Section VII concludes the study and outlines future research directions.
\section{Background and Literature Review}
Currently, there is a large amount of research and development in the safety paradigm of autonomous driving. To ensure a robust critical automotive safety system, two approaches are becoming popular: perception and map-generated infrastructure \cite{schafer2022fighting}.  The importance of a computer vision model in real-time perception for developing a framework for road object detection, especially to identify routes, pedestrians, and surrounding vehicles, is highlighted in \cite{uccar2017object}. Although vision-based systems work well in ideal settings, their performance often drops under poor lighting, fog, or heavy rain, which we address as drift situations. To address this, sensor fusion techniques that combine data from LiDAR, radar, and cameras have been introduced. Zhang et al. \cite{zhang2022robust} presented a deep fusion approach that significantly improved 3D object detection under harsh environmental conditions. Their work shows that achieving reliable perception depends on combining multiple sensor types and understanding the changing context. However, as AI models become more complex, understanding their decisions becomes more challenging. Explainable Artificial Intelligence (XAI) addresses this issue by providing insights into how and why certain decisions are made. Dong et al.\cite{dong2023did} proposed an XAI framework specifically designed for autonomous driving, revealing the reasoning behind actions such as braking or lane changes, crucial to increasing the transparency and trust of the system. In a related study, Atakishiyev et al.\cite{atakishiyev2024explainable} present a comprehensive overview of XAI techniques in autonomous driving, identifying key research directions and challenges. Their work emphasizes the importance of making AI decisions interpretable and highlights practical guidelines for implementing explainability in real-world driving systems.\\
Eclipse ADORE is a new addition to automotive research, which works as a modular toolkit for decision-making, motion planning, control, and simulation of automated vehicles. A foundational presentation by Heß et al. \cite{hess2022adore, maarssoe2025adore} outlines ADORe’s objectives: enabling research on cooperative automated vehicles (CAVs), infrastructure interaction, and multi-agent traffic systems. The authors in their papers stated that the core architecture of ADORE comprises decision-making modules for maneuver selection (e.g., lane-following, lane change, emergency stop), maneuver planning through optimization of longitudinal and lateral profiles with trajectory tracking. 
\section{Context Aware CV Model}
In our simulated autonomous driving system, we implemented YOLO (You Only Look Once) version 8 as the CV model that can manage drift scenarios at high speeds without compromising accuracy. The YOLO model is an object detection frameworks that excels at adapting to novel images and settings. Unlike other algorithms like R-CNN, SSD, it evaluates the entire image and acquires more generalized features during training. The goal of the computer vision model was to classify objects like road signs having different shapes and limits after being trained from the synthetic images generated from CARLA. In this study, CARLA is utilized to simulate vehicle dynamics and sensor inputs, enabling evaluation of the perception and control of the ADORe system under varied driving conditions.\\
Reliable perception is essential for the autonomous safety critical system, but traditional CV models often face data drift, where changes in input distributions caused by factors such as weather, construction zones, or sensor noise degrade detection accuracy. The goal of our context-aware CV model developed using YOLO was to mitigate such data drifts and ensure safety when the car is autonomous. Formally, data drift is a distributional shift that can be categorized into three main types:
\begin{itemize}
    \item \textbf{Covariate Drift}: Occurs when the input features change while the relationship between inputs and outputs remains stable (e.g., traffic signs appearing differently in foggy versus sunny weather).
    \item \textbf{Prior Probability Shift}: Happens when the frequency of labels changes while the input–label relationship stays constant (e.g., one speed limit sign class appearing far more often than others in deployment).
    \item \textbf{Concept Drift}: Arises when the relationship between inputs and labels itself changes over time (e.g. pedestrians crossing streets in new or less predictable patterns).
\end{itemize}
The developed context-aware perception model is capable to handle covariate and prior probability drift to improve object detection under real-world conditions. It enhances robustness through diverse lighting data and balanced class distributions. The proposed workflow ensures reliable perception outputs that support adaptive and secure decision making within the ADORE system.
\section{Integration of CARLA and ADORE}
To create a unified simulation and control framework, we integrated the CARLA simulator version 0.9.13 with the ADORE framework using the ROS (Robot Operating System) bridge version 1 NOETIC. This setup provides a middleware layer that allows real-time communication between the simulated environment and the decision-making modules. ADORE worked as a modular software platform that provides tools and libraries for perception, decision-making, control, and simulation of automated vehicles \cite{maarssoe2025adore}. ROS here functions as a middleware that facilitates communication between various software modules within autonomous systems. It is built on a decentralized, node-based architecture, where each node represents a process that performs a specific task, such as perception, control, or planning. At the core of the communication model is the publisher-subscriber mechanism, where data flows through named channels called topics. A node publishing data does not need to know who receives it and vice versa, allowing components to function independently while remaining interconnected. The exchange of messages is coordinated by the ROS Master, which acts as a directory service. When nodes are launched, they register their intention to publish or subscribe to specific topics. Although Master does not transmit data itself, it facilitates the discovery process, after which the nodes communicate directly in a peer-to-peer manner, ensuring low latency and scalable system performance \cite{quigley2009ros}. The \textbf{adore\_if\_carla} framework, developed by DLR's Transportation Systems (TS) department, was used as the integration tool and established through the \textit{carla\_ros\_bridge}, utilizing the Robot Operating System (ROS) as the communication middleware to enable data exchange and control.
\section{System Workflow}
\begin{figure}
    \centering
    \includegraphics[width=0.7\linewidth]{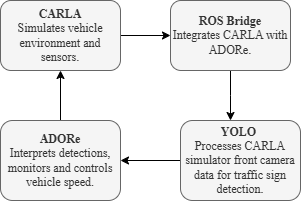}
    \caption{Functionalities of the Individual Modules in the Combined Workflow}
    \label{system_architecture}
\end{figure}

The system operates through a modular pipeline that allows real-time perception, decision making, and control within the CARLA simulator. RGB camera data from the ego vehicle is processed by a YOLOv8-based perception node to detect traffic signs or obstacles. The camera image frames are received from the CARLA simulator via an ROS image topic and converted into OpenCV format using the \texttt{ cv\_bridge} library. Each frame is passed through the YOLO model, which extracts object class IDs, confidence scores, and bounding boxes. After acquiring this information, the node publishes specific commands using the \texttt{AckermannDrive} message to ADORE, which then interprets these commands along with the current vehicle speed to determine the appropriate driving action, such as stopping, accelerating, or maintaining a certain speed limit. All commands are sent via ROS Bridge to adjust vehicle behavior accordingly, with dynamic acceleration constraints ensuring smooth motion. Communication between modules is handled using ROS topics, allowing real-time data exchange and integration between perception, decision, and control components. The control loop runs at 10 Hz, allowing smooth and responsive autonomous driving within the simulation environment. The overall workflow is illustrated in Figure~\ref{system_architecture}.
\section{System Verification}
\begin{figure}[h!] 
    \centering 
    \includegraphics[width=\linewidth]{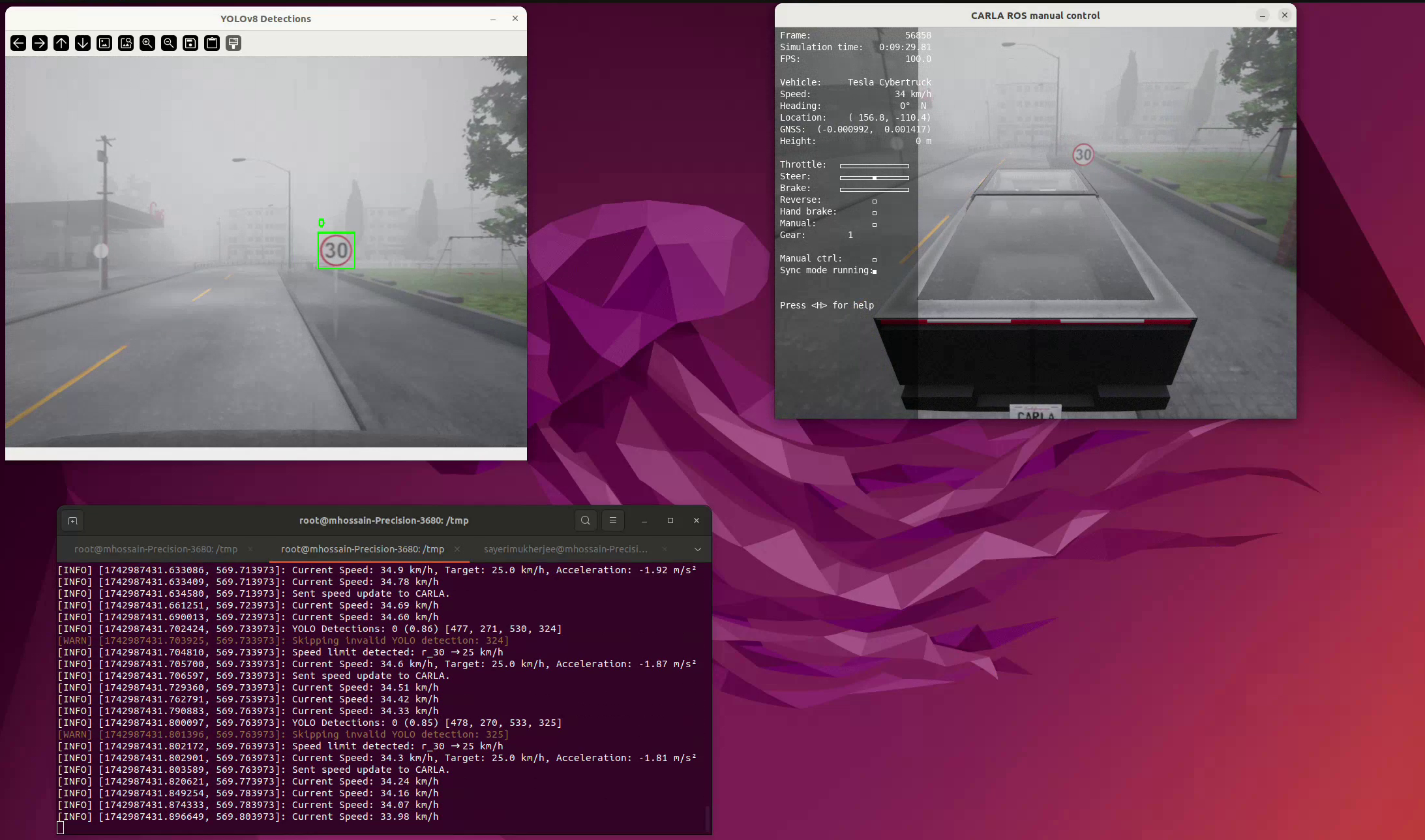} 
    \caption{Response of the Perception Model in Foggy Weather} 
    \label{fig:yolofog} 
\end{figure}

\begin{figure}[h!] 
    \centering 
    \includegraphics[width=\linewidth]{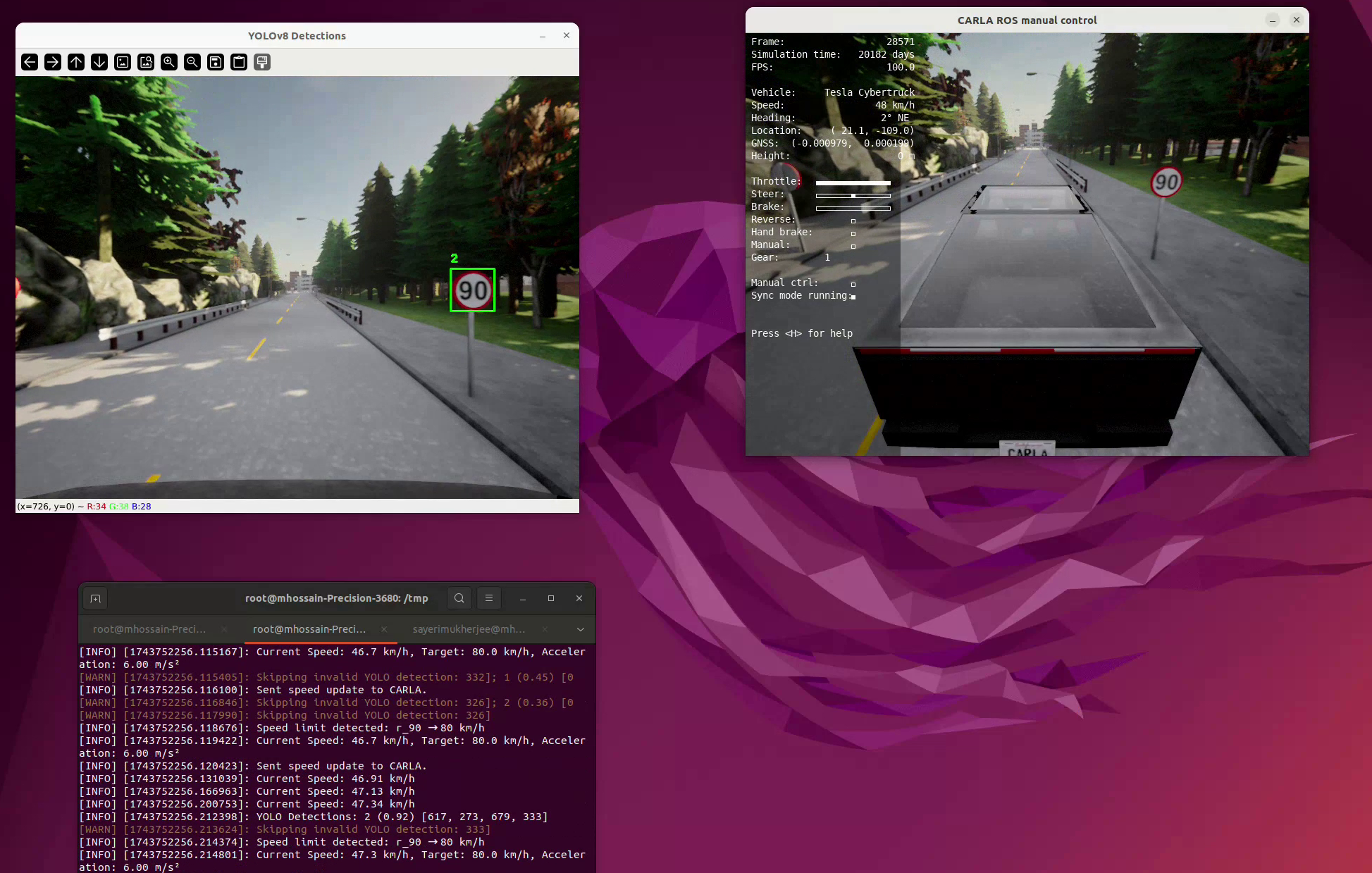} 
    \caption{Response of the Perception Model in Normal Weather} 
    \label{fig:yoloclear} 
\end{figure}

\begin{figure}[htbp]
    \centering
    \includegraphics[width=\linewidth]{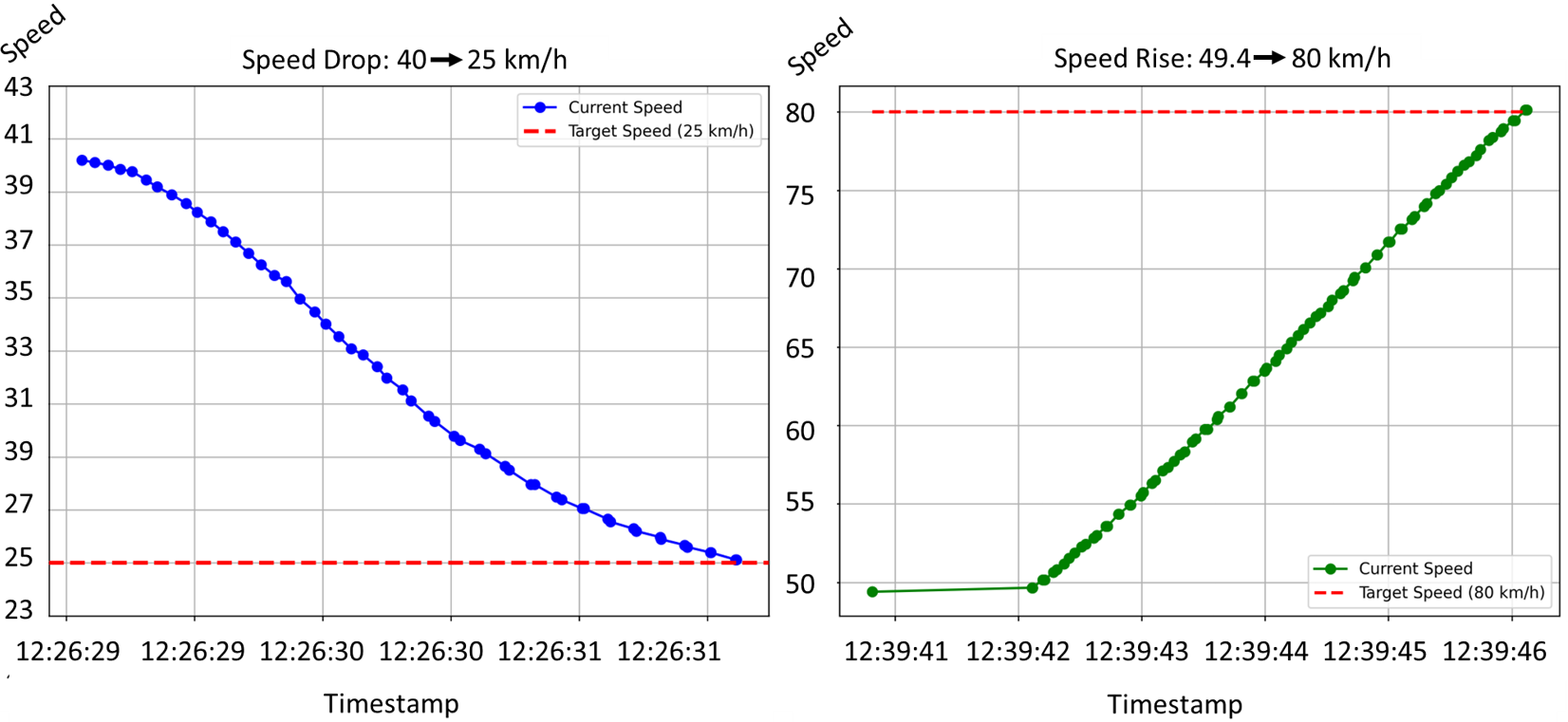}
    \caption{ADORE Response Time Following Perception Model}
    \label{fig:speed_profile}
\end{figure}

To verify the effectiveness of the system, we first measured the performance of the CV model in a similar way to that explained in our previous research \cite{hossain2024impact}. For training purposes, a selection of 4,296 images was made, while 767 images showcasing diverse road signs and weather conditions indicative of drift were used for validation. Performance of the model based on standards evaluation metrics can be found in Table \ref{tab:perf}. The model demonstrated a highly reliable and consistent performance in object detection tasks with a success rate of more than 90\% for all metrics. Its strong generalization and balanced precision recall characteristics indicate that it can effectively handle complex scenarios with high detection accuracy and robustness.
\begin{table}[!h]
\renewcommand{\arraystretch}{1.1}
\caption{\textsc{Validation Performance of CV Model}}
\label{tab:perf}
\centering
\begin{tabular}{|l|c|}
\hline
\textbf{Criteria} & \textbf{Validation} \\ \hline
Precision & 0.9827 \\ \hline
Recall & 0.9799   \\ \hline
F1-Score & 0.9812 \\ \hline
mAP50 & 0.9872 \\ \hline
mAP50-95 & 0.8700 \\ \hline
\end{tabular}
\end{table}
To further validate, we devised a test scenario to evaluate the runtime performance of the YOLO-based computer vision model and observe ADORE's reactions to it. In the test case, we simulated both clear and foggy weather with speed signs of 30km/h and 90km/h sequentially. Initially, the vehicle speed was set at 40km/h. The weather condition at the simulation runtime was set by publishing messages on the ROS topic \texttt{weather\_control}. This arrangement emulates scenarios of minimal lighting, poor visibility, and decreased solar altitude, adding visual intricacies that mirror actual conditions during foggy circumstances. It also helped to evaluate the robustness of the YOLO detector and the responsiveness of the ADORE controller in real time. Figures~\ref{fig:yoloclear} and~\ref{fig:yolofog} illustrate the detection performance of the YOLO model under clear and foggy weather, respectively. In both cases, traffic signs are accurately detected in real time, even under reduced visibility, demonstrating the robustness of the model to environmental variability.  The test case was executed on a local machine running Windows 11, equipped with 128GB of RAM. To facilitate CARLA operations, an NVIDIA RTX 6000 Ada Generation GPU with 48 GB of GDDR6 ECC memory was utilized.
Once a speed limit is identified, the ADORe node calculates the required acceleration or deceleration based on the difference between the current and target speeds. To ensure smooth and realistic driving behavior, the system limits maximum acceleration to 6.0 m/s² and deceleration to –6.0 m/s², while also applying a scaling factor of 0.7 to the speed error ($\Delta v$). This factor dampens the control response, avoiding sudden changes in velocity and enabling more stable and human-like motion. As shown in equation (\ref{eq:control}), the control law ensures proportional acceleration within safe limits. The system processes speed limit detections and updates vehicle commands with a measured response latency below 0.5 seconds, ensuring prompt and smooth adaptation to changing speed limits.
\begin{equation}
    a = \min \left( 6.0,\; 0.7 \times \Delta v \right)
    \label{eq:control}
\end{equation}
Figure~\ref{fig:speed_profile} illustrates the vehicle’s speed profile over time, highlighting the controller’s response to detected speed limit changes. In the first segment, the vehicle decelerates smoothly from 40.2~km/h to 25.0~km/h within approximately 2 seconds after successfully detecting a reduced speed limit (mapped from 30~km/h to 25~km/h) in foggy weather condition. In the next phase, the vehicle accelerates from 49.4~km/h to 80.0~km/h in under 5 seconds upon detecting an increased speed limit (mapped from 90~km/h to 80~km/h) in clear weather. Mapping to lower speed was done to ensure precise brake control without abrupt speed fluctuations and consistent throttle modulation. As the figures indicate, we observed smooth transitions in both phases, confirming the precise actuation and responsiveness of the unified system.
\vspace*{-0.2cm}
\section{Conclusion}
Our research shows that integrating deep learning-based perception with ADORE's adaptive control enhances autonomous driving safety in diverse conditions. Looking forward, this approach can be extended to address more complex navigation tasks and eventually deployed in real vehicles, as ADORE is designed for integration into real-world automotive systems.
\section*{Acknowledgment}
This work has been funded by 1) the Federal Ministry of Education and Research (BMBF) as part of AutoDevSafeOps (01IS22087Q) and 2) the research project EcoMobility through the European Commission (101112306) and BMBF (16MEE0316).
\bibliographystyle{IEEEtran}
\bibliography{main}
\end{document}